%% file: arxiv.tex
\documentclass{article}
\usepackage[accepted]{aistats2018} 

\usepackage{times}
\usepackage{graphicx} \usepackage{subcaption}

\usepackage[numbers]{natbib}

\usepackage{algorithm}
\usepackage{algorithmic}

\usepackage{hyperref}

\usepackage[T1]{fontenc}    
\usepackage{url}            
\usepackage{booktabs}       
\usepackage{amsfonts}       
\usepackage{nicefrac}       
\usepackage{microtype}     
 \usepackage{array}
\usepackage{pbox}

\usepackage{times}
\usepackage{mathtools}

\usepackage{color}
\usepackage{multirow}
\usepackage{amsthm}
\usepackage{mathrsfs}
\usepackage{graphicx}
\usepackage{caption}\usepackage{xspace}
\usepackage{float}
\usepackage{wrapfig}
\usepackage{enumitem}
\setitemize{noitemsep,topsep=0pt,parsep=2pt,partopsep=0pt}

\usepackage{multibib} \newcites{sup}{Supplementary References}

\input{notation.tex}

\input{theorem_notation.tex}

\DeclareMathOperator{\cone}{cone}
\DeclareMathOperator*{\argmin}{arg\,min}

\DeclareMathOperator{\conv}{conv}
\DeclareMathOperator{\faces}{faces}

\DeclareMathOperator{\diam}{\mathrm{diam}}
\DeclareMathOperator{\radius}{\mathrm{radius}}

\DeclareMathOperator{\proj}{proj}

\newcommand{\lmo}{\textsc{LMO}\xspace}

\newcommand{\Cf}{C_{f,\cA}}

\newcommand{\dkl}{D^{KL}}

\begin{document}

\twocolumn[
\aistatstitle{Boosting Variational Inference: an Optimization Perspective}

\aistatsauthor{
  Francesco Locatello
  \And
  Rajiv Khanna
  \And
  Joydeep Ghosh
  \And 
  Gunnar R{\"a}tsch
  }

\aistatsaddress{
MPI for Intelligent Systems \\ ETH Zurich
  \And
  UT Austin
  \And
  UT Austin
  \And 
  ETH Zurich
  }
]

\begin{abstract}
Variational inference is a popular technique to approximate a possibly intractable Bayesian posterior with a more tractable one. Recently, boosting variational inference~\cite{Miller:2016vt,Guo:2016tg} has been proposed as a new paradigm to approximate the posterior by a mixture of densities by greedily adding components to the mixture. However, as is the case with many other variational inference algorithms, its theoretical properties have not been studied. In the present work, we study the convergence properties of this approach from a modern optimization viewpoint by establishing connections to the classic Frank-Wolfe algorithm. Our analyses yields novel theoretical insights regarding the sufficient conditions for convergence, explicit rates, and algorithmic simplifications. Since a lot of focus in previous works for variational inference has been on tractability, our work is especially important as a much needed attempt to bridge the gap between probabilistic models and their corresponding theoretical properties.
\end{abstract}
\section{Introduction}

Variational inference~\cite{Blei:2016vr} is a method to approximate complicated probability distributions with simpler ones. In many applications, calculating the exact posterior distribution is intractable, and methods like MCMC while being flexible can also be prohibitively expensive. Variational inference restricts the posterior to be a member of a simpler and more tractable family of distributions, and the inference problem reduces to finding this member that can ``closely'' represent the true underlying posterior. The closeness is typically measured in the KL sense. 

One of the most commonly used family of distributions for the tractable set is the so called \textit{mean field family}, which assumes a factored structure. An example of such a family is the set of Gaussian distributions with diagonal covariance matrices. While the inference is computationally efficient due to the properties of Gaussian distributions, this family can be too restrictive. As such, the approximated distribution is often not a good representation of the true posterior. A simple example is a multi-modal distribution. The mean field family will be able to only capture one of the modes. 

There have been a number of efforts to improve the approximation while retaining the simplicity of Gaussian distributions. For example, one could consider approximating by a mixture of Gaussian distributions and allowing more than just isotropic structures. A mixture of isotropic Gaussian distributions is already a much more powerful and flexible model than a single isotropic Gaussian. In fact, it is flexible enough to model any distribution arbitrarily closely~\cite{parzen1962estimation}. While there has been significant algorithmic  and empirical development for studying variational inference using mixture models~\cite{Miller:2016vt,Guo:2016tg,Li:2000vt,Li1999thesis}, there have been limited theoretical studies. In this work, our aim is to bridge this gap. 

We study, from an optimization perspective, the approximation of a posterior by iteratively adding simpler distributions, not necessarily Gaussians, \emph{greedily}~\cite{Guo:2016tg}. 
Given that one can find the components of the mixtures, building a mixture is a convex problem which we show have efficient algorithms converging to the global optimum. On the other hand, finding these individual components is non-convex and is known to exhibit several local optima \cite{mandt2016variational,Blei:2016vr}. However, we show that one does not need to solve the inner non-convex problem exactly to achieve the same strong convergence guarantees. The key to our analyses is establishing connections with a functional variant of the well known Frank-Wolfe Algorithm~\cite{jaggi13FW}. This connection helps us provide the convergence rate of the greedy variational boosting algorithm with explicit constants in terms of the properties of the  distributions. 

To the best of our knowledge, these explicit rates have not been known before in the context of variational inference. Moreover, we are also able to provide novel insights, including sufficient conditions for a linear convergence as opposed to the previously conjectured sublinear $\mathcal{O}(1/T)$ rates where $T$ is the number of iterations. Our contributions are both algorithmic and theoretical:
\begin{itemize}
 \setlength\itemsep{-0.5em}
\item We connect boosting variational inference (Algorithm~2 in \cite{Guo:2016tg}) with the Frank-Wolfe framework \cite{Jaggi:2013wg} enabling us to carefully analyze its convergence. We also thoroughly analyze the assumptions essential to ensure global convergence and present an explicit rate (with constants) for their conjectured $\mathcal{O}(1/T)$ rate.\\
\item We propose simpler variants of the same algorithm that retain the same strong theoretical properties (fixed step size and closed-form line search in Algorithm~\ref{algo:FW}).\\
\item We provide sufficient conditions under which greedy algorithms achieve linear $(\mathcal{O}(e^{-T}))$ convergence and therefore are much faster than what was previously conjectured.\\
\item We revisit the Norm-Corrective Frank-Wolfe in Algorithm~\ref{algo:normCorrectiveFW} and give linear convergence guarantees at the cost of a slightly larger computational cost. This algorithm allows one to selectively reoptimize all the weights of the mixture efficiently at every iteration resulting in much faster convergence in practice.
\end{itemize}

\subsection{Related work}
Variational approximations by using mixture models has been extensively studied and applied. Perhaps the closest algorithmic setup to our work is that of~\cite{Guo:2016tg}. They iteratively add components to the mixture greedily, similar to gradient boosting. They require the boosting subroutine to return the optimal density but as we show, this is not required for obtaining their conjectured convergence rate of $O(1/T)$, where $T$ is the number of components added.~\cite{Miller:2016vt} also use a very similar algorithm in their setup.  

Traditional approaches directly target the non-convex problem of finding exactly the first density of the mixture. For this problem, some convergence analysis was carried out by~\cite{khan2015faster}, but their rates are only applicable \emph{locally}, as they depend on a smoothness assumption of the KL divergence which does not hold globally unless the iterate is close to the optimum~\cite{nielsen2009statistical}. As we will see, greedy methods have the clear advantage that one does not need to perfectly find the best approximating distribution in the family as previously considered by~\cite{Guo:2016tg}. A rough approximate solution is enough to ensure convergence.

The Frank-Wolfe Algorithm~\cite{jaggi13FW} is a popular algorithm for convex constrained minimization, and is specially attractive because of its cheap projection-free iterations. The algorithm is well studied both theoretically and empirically~\cite{LacosteJulien:2015wj,TangECCV14,jain2016structured}, and has even been applied to non-euclidean spaces. For example,~\cite{KrishnanLS15} consider a variational objective for approximate marginal inference over the marginal polytope. 

The rest of the paper is organized as follows. We review the variational inference problem from an optimization perspective in Section~\ref{sec:reviewVI} and the necessary and sufficient assumptions that are required to show convergence in Section~\ref{sec:domainRestrict}. We present our further algorithmic contributions for the framework in Section~\ref{sec:FW}.  We conclude the paper with an experimental proof of concept showing that the proposed methods converge as expected.

\paragraph{Notation.} 
We represent vectors by small letters bold, e.g. $\bx$ and matrices by capital bold, e.g., $\bX$. 
For a non-empty subset $\cA$ of some Hilbert space $\cH$, let $\conv(\cA)$ denote its convex hull. $\cA$ is often called \textit{atom set} in the literature, and its elements are called \textit{atoms}. Given a closed set $\cA$, we call its diameter $\diam(\cA)=\max_{\bz_1,\bz_2\in\cA}\|\bz_1-\bz_2\|$ and its radius $\radius(\cA) = \max_{\bz\in\cA}\|\bz\|$.
The support of a density function $q$ is a measurable set denoted by capital letters sans serif i.e. $\sfZ$. Sometimes, we write the domain of a density function with the same notation, but if the domain and the support do not coincide it would be made explicit.
The inner product between two density functions $p,q:\sfZ\rightarrow\bbR$ in $L^2$ is defined as $\langle p, q\rangle:=\int_{\sfZ}p(z)q(z)dz$. 
\vspace{-2mm}
\section{Variational Inference Problem Setting}\label{sec:reviewVI}
\vspace{-2mm}
Say, we observe $N$ data points $\bx$ from some space. The Bayesian modelling approach consists of specifying a prior $\pi(\bz)$ on the data and the likelihood $p(\bx|\bz)$ for some parameter vector $\bz\in \sfZ$ where $\sfZ$ is a measurable set, for example $\bbR^D$ \cite{Blei:2016vr}. One of the challenges of Bayesian inference is that the posterior, obtained through Bayes theorem could be intractable because of a hard to calculate normalization constant. Instead, the joint distribution is usually easier to evaluate i.e. $p(\bx,\bz)$. From a functional perspective, the posterior can be written as $p_{\bx}(\bz):\sfZ\rightarrow \bbR^+_{>0}$. We assume that $p_{\bx}(\bz)\neq 0 \ \forall \bz\in\sfZ$. We use $p_\bx$ to represent the posterior and $p$ for the joint distribution.
The goal of variational inference is to find a density from a constrained set of tractable densities $\cQ$ with support $\sfQ$, $q:\sfQ\rightarrow (0,\infty), q\in\cQ$ that is close in the KL sense to the true posterior. The respective optimization problem is:
\begin{align}\label{eq:VarInfProb}
\min_{q\in\cQ} \dkl(q\|p_{\bx}).
\end{align}
Note that an unconstrained minimization would yield $q$ to be equal to the true posterior. Thus, one would ideally want the set $\cQ$ to be able to represent the parameter space $\sfZ$ well, while still retaining tractability. The objective in Equation~\eqref{eq:VarInfProb} is not computable as it requires access to $p_{\bx}(\bz)$~\cite{Blei:2016vr}. Instead, it is common practice to maximize the so called the evidence lower bound (ELBO), given by:
\begin{equation}
-\bbE \left[ \log q(\bz)\right] + \bbE \left[ \log p(\bx,\bz)\right]
\end{equation}
It is easy to see that maximizing the ELBO, is equivalent to solving the following optimization problem:
\begin{equation}\label{eq:minklQ}
\min_{q\in\cQ}\dkl(q||p)
\end{equation}
While it is well known that $\dkl$ is strictly convex in $q$, its smoothness and strong convexity depends on the choice of $\cQ$. \cite{wang2016boosting,Guo:2016tg} showed that the smoothness constant can be bounded by the minimal value obtained by all pdf functions of the densities in $\cQ$ in their domain and \cite{wang2016boosting} showed that the strong convexity constant is equal to the respective maximal value.
\vspace{-2mm}
\section{Domain Restricted Densities for Variational Inference}\label{sec:domainRestrict}
\vspace{-2mm}
For simplicity in the following we write $\dkl(q)$ instead of $\dkl(q||p_{\bx})$.
A sufficient condition for smoothness of the $\dkl(q)$ is that the density $q$ is bounded away from zero~\cite{Guo:2016tg}. We extend this result, showing the necessary condition for global smoothness of $\dkl(q)$ to hold on the entire support $\sfQ$. 
\begin{lemma}\label{lemma:smooth}
$\dkl(q)$ is Lipschitz smooth on $\cQ$ with constant $L = \frac1\epsilon$ if and only if $q/p_{\bx}:\sfQ\rightarrow[\epsilon,\infty)$ with $\epsilon>0$ i.e. is bounded away from zero in $\sfQ$. A sufficient condition for smoothness of $\dkl(q)$ is $q:\sfQ\rightarrow[\epsilon,\infty)$ with $\epsilon>0$ i.e. is bounded away from zero in $\sfQ$. \end{lemma}
 Smoothness is a typical assumption which is useful to measure the convergence of optimization algorithms and was employed also in the variational inference setting \cite{khan2015faster}. Lemma~\ref{lemma:smooth} entails that the proofs based on smoothness are valid only in some regions of the space. 
 
Lemma~\ref{lemma:smooth} states that if $q$  is a good approximation for $p_{\bx}$  (i.e. their ratio is bounded away from zero) then $\dkl$ is smooth. If one consider a general density $q$, a simple way to ensure smoothness is to bound $q$ away from zero. Therefore, we restrict the support of the approximating densities to compact sets. In practice, if the algorithms are initialized well enough, $q/p_{\bx}$ can be bounded away from zero. 
As an example, consider a mixture of two Gaussians with mean in $\bbR^1$ sufficiently far apart. The boosting approach place a density on one of the modes first and then to the other. Therefore, the gradient of the $\dkl$ at the second iteration -- $\log(q_1/p_{\bx})$ -- is arbitrarily large in some parts of the domain depending on how far are the modes and the covariance matrix of $q_1$. Unfortunately, those are precisely the parts the method targets. Thus, we need to ensure that a significant mass is placed on the second mode as well. For a family of densities which is not bounded away from zero, truncating the support can be seen as a smoothing condition. Initializing with the solution of the mean field variational inference would place some mass on both the modes, so the $\dkl$ would appear smooth to the algorithm and truncation might not be necessary. While this is valid in practice, we focus on truncated densities as we need to ensure that the rates we present in this work are valid for any density in the set $\cA$ independently of $p_{\bx}$ and any initial approximation. 
Following the line of work of \cite{koyejo2014prior,khanna2015sparse} we introduce the information projection from $\cQ$ to another set $\cA$ where all the densities $q\in\cA$ are obtained by truncating densities from $\cQ$ and therefore have bounded support $\sfA$. 
Intuitively, variational inference aims at projecting the true posterior on the set of tractable densities $\cQ$ (for example factorial in the mean field case). 
Instead, the boosting variational inference considers mixtures of densities from the set $\cQ$, i.e., the optimization is constrained to $\conv(\cQ)$. The underlying intuition is that $\conv(\cQ)$ is more expressive than $\cQ$. For example, any density can be approximated with a mixture of Gaussian distributions with some appropriate covariance matrix.
In order to comment about the rates of convergence, we further restrict the densities in $\cQ$ to have a truncated support $\sfA\subseteq\sfQ$ and we call this set $\cA$. Therefore, $q(\bz):\sfA\rightarrow[\epsilon,\infty)$ with $\epsilon>0$ and $q(\bz)=0 \ \forall \bz\in\sfQ\setminus\sfA$. 
To distinguish a density in $\cQ$ and its truncated version in $\cA$ we write $q_\sfQ\in\cQ$ for the former and $q\in\cA$ for the latter. 

Therefore, we solve the following optimization problem:
\begin{align}\label{eq:BoostingVi}
\argmin_{\substack{q\in\conv(\cA)}}\dkl(q||p_{\bx}).
\end{align}
As  the original posterior $p_\bx$ has support $\sfZ$, the choice of $\conv(\cA)$ as optimization domain is suboptimal wrt $\cQ$ or $\conv(\cQ)$ as its support is a subset $\sfA\subseteq\sfQ\subseteq\sfZ$. We now measure exactly the error which is introduced truncating the support.

Let us first consider the projection of $p_\bx$ onto $\sfA$ (i.e. restrict the support of $p$ from $\sfZ$ to $\sfA$). We then have that:
\begin{align*}
p_{\sfA}(\bz) =  \begin{cases} \frac{p_{\bx}(\bz)}{\int_\sfZ p_{\bx}(\bz)\delta_\sfA(\bz)d\bz}, & \mbox{if } \bz\in\sfA \\ 0, & \mbox{otherwise }\end{cases}
\end{align*}
Where $\delta_\sfA(\bz)$ is the delta set function. Using the definition of $p_{\sfA}(\bz)$ we have that:
\begin{align}\label{eq:klstar}
\dkl(p_\sfA||p_{\bx}) &= \int_\sfA p_\sfA \log\frac{p_\sfA}{p_{\bx}} d\bz \nonumber\\
 &= \int_\sfA p_\sfA \log\frac{p_{\bx}}{p_{\bx}\cdot p_{\sfZ}(\bz\in\sfA) } d\bz \nonumber \\
 &=  -\log p_{\sfZ}(\bz\in\sfA)
\end{align}
This error represent a tradeoff between the smoothness of the objective (and therefore the rate of the boosting algorithm) and the quality of the approximation. 
The hope, is that $\conv(\cA)$ is a richer family of distributions than $\cQ$ (i.e. mean field variational inference) and is more tractable than both $\cQ$ and $\conv(\cQ)$ from the optimization perspective. Note that $p_\sfA$ does not have to be in $\conv(\cA)$. If $\cA$ contains non-degenerate truncated Gaussian distributions with some appropriate covariance matrix then $\conv(\cA)$ contains $p_\sfA$ which becomes the minimizer $q^\star$ of Equation~\eqref{eq:BoostingVi}.

In the rest of the paper, we consider the set $\cA$ as the set of non degenerate truncated distributions (upper and lower bound on the determinant of the covariance matrix). 
We assume that the elements in $\cA$ have all the following:
\begin{itemize}
\item[A1.] truncated densities with bounded support $\sfA$\\
\item[A2.] $q(\bz)\geq \epsilon>0 \ \forall \ \bz\in\sfA$ and $q$ is bounded from above by $M$\\
\end{itemize} 

Under these assumption, we can analyze some of the properties of the optimization domain. 
 
\begin{theorem}\label{thm:mfBounded}
The set $\cA$ of non degenerate truncated distributions bounded from above and compact support $\sfA$ is a compact subset of $\cH$.
\end{theorem}
The proof is deferred to the Appendix~\ref{app:proofs}.
Due to the convenient form of $\cA$ we can also compute its diameter as:
\begin{corollary}
Given a distribution $q\in\cA$, it holds that $\diam(\cA)^2\leq \max_{q\in\cA}4\|q\|^2\leq 4 M^2 \cL(\sfA)
$ where $\cL(\sfA)$ is the Lebesgue measure of the support $\sfA$, which is bounded under the assumptions of Theorem~\ref{thm:mfBounded}.  
\end{corollary}
We will extensively discuss the impact of these assumptions on both the convergence and the approximation quality in Section~\ref{sec:FW}.
\vspace{-2mm}
\section{Functional Frank-Wolfe For Density Functions}\label{sec:FW}
\vspace{-2mm}
In this section, we explain the foundations of boosting via Frank-Wolfe in function spaces. In the analysis of \cite{Wang:2015uw}, the authors enforce a bounded polytope using functions in $L^1$ with bounded $L_\infty$ norm. Following the more traditional approaches of \cite{Jaggi:2013wg,LacosteJulien:2015wj,Locatello:2017tq}, we further assume that the functions must have bounded $L_2$ norm. 

The optimization problem we want to solve is:
\begin{equation}
\min_{q\in\conv(\cA)}f(q).
\end{equation}
where $\cA\subset L^2$ is compact (see Theorem~\ref{thm:mfBounded}) and $f$ is a convex functional over $\conv(\cA)$ with bounded curvature over the same domain. The curvature is defined as in \cite{Jaggi:2013wg}:
\begin{equation}
\label{def:Cf}
\Cf := \sup_{\substack{s\in\cA,\, q \in \conv(\cA)
 \\ \gamma \in [0,1]\\ y = q + \gamma(s- q)}} \frac{2}{\gamma^2} D(y,q),
\end{equation}
where 
\begin{equation*}
D(y,q) :=f(y) - f(q)- \langle y -q, \nabla f(q)\rangle.
\end{equation*}
It is known that $\Cf\leq L\diam(\cA)^2$ if $f$ is $L$-smooth over $\conv(\cA)$. Due to Lemma~\ref{lemma:smooth}, we know that the $\dkl(q)$ with $q\in\cA$ is smooth which implies that the curvature is bounded. Therefore, $\dkl(q)$ is a valid objective for the FW framework.
In each iteration, the FW algorithm queries a so-called linear minimization oracle ($\lmo$) which solves the optimization problem:
\begin{align}\label{eq:lmo}
\lmo_{\cA}(y) := \argmin_{s\in\cA}\langle y,s\rangle
\end{align}
for a given $y\in\cH$ and $\cA\subset \cH$. As computing an exact solution of \eqref{eq:lmo}, depending on $\cA$, is often hard in practice, it is desirable to rely on an approximate $\lmo$ that returns an approximate minimizer $\tilde s$ of \eqref{eq:lmo} for some accuracy parameter $\delta$ and the current iterate $q^t$ such that:
\begin{align}
\langle y,\tilde s-q^t\rangle\leq \delta\min_{s\in\cA}\langle y,s-q^t\rangle
\end{align}
The $\lmo$ is, in general, a hard optimization problem. Therefore, an approximate solution is commonly employed.
We discuss a simple algorithm to implement the $\lmo$ in Section~\ref{sec:LMO}.
The Frank-Wolfe algorithm is depicted in Algorithm~\ref{algo:FW}. Note that Algorithm 2 in \cite{Guo:2016tg} is a variant of Algorithm~\ref{algo:FW}.

\begin{algorithm}[h]
\caption{Affine Invariant Frank-Wolfe}
  \label{algo:FW}
\begin{algorithmic}[1]
  \STATE \textbf{init} $q^{0} \in \conv(\cA)$  \STATE \textbf{for} {$t=0\dots T$}
  \STATE \qquad Find $s^t := (\text{Approx-}) \lmo_\cA(\nabla f(q^{t}))$
    \\  \STATE \qquad Variant 0:  $\gamma=\frac{2}{t+2}$\\
   \STATE \qquad Variant 1: $\gamma =\min\left\lbrace 1, \frac{\langle-\nabla f(q^t),s^t-q^t\rangle}{\Cf}\right\rbrace$
   \STATE \qquad Update $q^{t+1}:= (1-\gamma)q^{t} + \gamma s^t$\\
 \STATE \textbf{end for}
\end{algorithmic}
\end{algorithm} 

Algorithm~\ref{algo:FW} is known to converge sublinearly with the following rate.
\begin{theorem}[\cite{Jaggi:2013wg}]
\label{thm:sublinearFWAffineInvariant}
Let $\cA \subset \cH$ be a compact set and let $f \colon \cH \to \R$ be a convex function with bounded curvature $\Cf$ over~$\cA$.
Then, the Affine Invariant Frank-Wolfe algorithm (Algorithm~\ref{algo:FW}) converges for $t \geq 0$ as\vspace{-1mm}
\begin{equation*}
f(q^t) - f(q^\star) \leq \frac{2\left(\frac1\delta \Cf+\varepsilon_0\right)}{\delta t+2}
\end{equation*}
where $\varepsilon_0 := f(q^0) - f(q^\star)$ is the initial error in objective, and $\delta \in (0,1]$ is the accuracy parameter of the employed approximate \lmo.
\end{theorem}

In some cases convergence might actually be faster (i.e. linear), as stated below.
\begin{theorem}[\cite{guelat1986some}]\label{thm:linearInterior}
Let $\cA \subset \cH$ be a compact set and let $f \colon \cH \to \R$ be a strongly convex function with bounded curvature $\Cf$ over~$\cA$. Further, assume $q^\star$ lies within relative interior of $\conv(\cA)$.
Then, the Affine Invariant Frank-Wolfe algorithm (Algorithm~\ref{algo:FW}) produces a sequence of iterates that converges goemetrically to $q^\star$
\end{theorem}

\paragraph{Discussion:}
Recall that $\Cf\leq L\diam(\cA)^2$. In Theorem~\ref{thm:mfBounded} we showed that the set of non degenerate truncated distributions is bounded and in Lemma~\ref{lemma:smooth} we showed that the $\dkl$ exhibits bounded curvature on $\cA$. These results are important as they theoretically justify why we can successfully build a mixture of distributions approximating the posterior in a boosting-like approach. These optimization subtleties were not addressed in \cite{Guo:2016tg,Miller:2016vt} but are essential for the convergence of Algorithm~\ref{algo:FW}. In Theorem~\ref{thm:linearInterior} we introduce the idea that greedily adding a density in a boosting fashion is converging linearly under some additional assumptions. As one can not check whether the optimum is in the relative interior or not, we now focus on the sublinear rate, trying to understand how the assumptions which are made on the target family of distributions influence the convergence.  

We now characterize the constants in Theorem~\ref{thm:sublinearFWAffineInvariant} for the boosting variational inference problem.
\begin{theorem}\label{thm:explicitConstants}
Let the set $\cA$ satisfy A1 and A2. Then, it holds that:
\begin{align*}
\Cf\leq L\diam(\cA)^2\leq 4\frac{M^2} {\epsilon}\cL(\sfA)
\end{align*} 
\end{theorem}
\begin{corollary}
\label{thm:explicitRate}
Under the assumption of Theorem~\ref{thm:explicitConstants}, the Affine Invariant Frank-Wolfe algorithm (Algorithm~\ref{algo:FW}) converges for $t \geq 0$ as
\begin{equation*}
f(q^t) - f(q^\star) \leq 8\frac{M^2 \cL(\sfA)}{\epsilon(\delta^2 t+2)} + \frac{2\varepsilon_0}{\delta t+2}
\end{equation*}
where $\varepsilon_0 := f(q^0) - f(q^\star)$ is the initial error in objective, and $\delta \in (0,1]$ is the accuracy parameter of the employed approximate \lmo.
\end{corollary}
\paragraph{Discussion:}
As expected, the rate depends on the two main assumptions we introduced: compact support and non degenerate distributions. The support and covariance matrix directly influence the values of $\epsilon$ and $M$. This is substantially different to what is presented in \cite{Guo:2016tg,Miller:2016vt} as the explicit assumptions we make allows us to understand how the choices in the distribution family influences the rate. In particular, \cite{Guo:2016tg,Miller:2016vt} did not consider the importance of bounded supports, and as we show that it is vital for their conjecture of $O(1/t)$ to hold. Similarly, the sublinear convergence analysis of variational inference of \cite{khan2015faster} only holds where the ratio $q/p_{\bx}$ is bounded (recall from Lemma~\ref{lemma:smooth}).

If the set $\cA$ contains truncated Gaussian distributions with non-degenerate covariance matrix but with small enough determinant to perfectly approximate any density defined on a bounded support it also satisfies A1 and A2.
We can now write the suboptimality of the boosting approach, making the tradeoff between the support and the approximation error in term of $\dkl$ explicit. Indeed, in Equation~\eqref{eq:klstar} we compute the information lost in the projection on a compact support. On the other hand, $q^\star$ represent the projection of $p$ onto the support $\sfA$ as well. Therefore, we can finally give the Theorem that measures the total information loss of boosting variational inference via Frank-Wolfe.
\begin{theorem}\label{thm:FWsublinearGaussian}
Let the set $\cA$ of non degenerate truncated Gaussian distribution have compact support $\sfA\in\bbR^d$. Further assume that their means are in $\sfA$ and their covariance matrix before truncation is given by $\sigma^2\bI$ with $\sigma\geq\sigma_{\min}>0$ with $\sigma_{\min}$ being small enough such that $p_{\sfA}\in\conv(\cA)$. Let $\ba$ and $\bb$ be the vertices of the diameter of $\sfA$. Then, the information loss of the Affine Invariant Frank-Wolfe algorithm (Algorithm~\ref{algo:FW}) with some choice of the compact support $\sfA$ converges for $t \geq 0$ as\vspace{-1mm}
\begin{align*}
\dkl(q^t||p) &\leq \frac{4P(\cN(\ba,\sigma_{\min}^2\bI)\in\sfA)}{\sigma_{\min}^{\frac{d}{2}}2^{\frac{d}{2}} K^2}\exp\left(\frac12\frac{\diam(\sfA)^2}{\sigma^2_{\min}}\right)\\
&\qquad\frac{1}{\delta^2 t+2} + \frac{2\varepsilon_0}{\delta t+2} -\log p(\bz_{\sfZ\setminus\sfA} = 0)
\end{align*}
where $\varepsilon_0 = \dkl(q^0||p) - \dkl(q^\star||p)$, $\delta \in (0,1]$ is the accuracy parameter of the employed approximate \lmo,  $p$ is the true posterior distribution and $K := min_{\bmu\in\sfA}P(\cN(\bz,\bmu,\sigma_{max}^2I)\in\cA)$. Note that $K$ is bounded away from zero.
\end{theorem}
\paragraph{Discussion:}
Note that the diameter of $\cA$ is related to the L2 norm of its elements. If the dimensionality increases, this notion of distance loses meaning (curse of dimensionality). This explicit dependency in the rate is an artifact of the proof technique as a consequence of using the L2 norm. Note that $K$ depends implicitly on $d$ and it decreases whenever $d$ increases and the support of $\cA$ remains fixed~\cite{hopcroft2014foundations}.
Understanding whether the rate is meaningful in high dimensions is a challenging question. Better rates might be achieved with a different notion of distance and are left as future work.

\subsection{Implementing the LMO}\label{sec:LMO}
To solve the $\lmo$ problem we revisit a technique well known in the stochastic variational inference framework \cite{ranganath2014black, khan2015faster} to account for our constrained scenario.
Let us rewrite the optimization problem of Equation~\eqref{eq:lmo} exploiting the parametric form of the distributions in $\cA$ as:
\begin{align*}
\argmin_{\theta:s(\theta)\in\cA} \langle s(\theta),\nabla f(q^t)\rangle = \argmin_{\theta:s(\theta)\in\cA} \bbE_{\bz\sim s(\theta)}\left[\nabla f(q^t(\bz))\right]
\end{align*}
In order to obtain a valid solution of the $\lmo$ problem, we perform projected gradient descent on the parameters of $s(\bz;\theta)$ with a stochastic approximation of the gradient. Let $\proj_{\cA}$ be an operator such that $proj_{\sfA}\left[s(\bz)\right]\in\cA$ holds. 
This operator is easy to implement in the Gaussian case, as it is reduced to a box constraint for the mean, a constraint on the eigenvalues of the covariance matrix and a truncation. 
We therefore sample $S$ points from $s(\bz;\theta)$ and use the following estimator for the gradient:
\begin{align}
\nabla_\theta \bbE_{\bz\sim s(\bz;\theta)}\left[\nabla f(q^t(\bz))\right] &= \int_{\cD}\nabla f(q^t(\bz))\nabla_{\theta}s(\bz;\theta)d\bz\nonumber\\
&= \int_{\cD}\nabla f(q^t(\bz))s(\bz;\theta)\cdot\nonumber\\
&\quad\quad\nabla_{\theta}\log s(\bz;\theta)d\bz\nonumber\\
&\approx \frac1S\sum_{s=1}^S \nabla f(q^t(\bz^{(s)}))\cdot\nonumber\\
&\quad\quad\nabla_\theta \log s(\bz^{(s)};\theta)\nonumber\\
&=:\hat\nabla_\theta \bbE_{\bz\sim s(\bz;\theta)}\left[\nabla f(q^t(\bz))\right]\label{eq:stoch_grad}
\end{align}
where the $\bz^{(s)}$ are sampled from $s(\bz;\theta)$.
This stochastic approximation of the gradient is known to suffer from high variance. Any of the known variance reduction techniques known can be used, e.g., see \cite{ranganath2014black}. 

We now perform a projected gradient step as: 
\begin{align}
s^{l+1}(\bz;\theta) &= proj_{\cA}\left[s^{l}(\bz;\theta)\right. \nonumber\\
&\quad\left. - \eta \cdot  \hat\nabla_\theta \bbE_{\bz\sim s^{l}(\bz;\theta)}\left[\nabla f(q^t(\bz))\right]\right]\label{eq:proj_grad_step}
\end{align}
 for some stepsize $\eta$. Note that $\hat\nabla_\theta \bbE_{\bz\sim s^{l}(\bz;\theta)}\left[\nabla f(q^t(\bz))\right]$ is an unbiased estimator for the gradient as showed in \cite{l1995note}. Further approximation is possible in the data domain as the sampling process is i.i.d. and $\nabla f(q^t)= \log\frac{q^t(\bz)}{p(\bx,\bz)}$.
The stochastic LMO algorithm is depicted in Algorithm~\ref{algo:LMOstoc}. 
Notably, an approximate solution of the LMO is sufficient to ensure convergence, even if it is $\delta$-approximate only in expectation \cite{Jaggi:2013wg}. Therefore, relying on cheap estimates of the gradient is well posed in this framework. 

Note that the linear problem of Equation~\eqref{eq:lmo} without the constraints would be trivially solved by a degenerate distribution placed on the minimum value of the gradient. Therefore, if the set $\cA$ contains truncated normal distributions there is a local minimum with covariance $\sigma_{\min}\bI$. Therefore, in the experiments we do not learn the covariance matrix. Recall that an approximate solution for the LMO problem  is enough to converge.
\begin{algorithm}
\begin{algorithmic}[1]
  \STATE \textbf{init} $s^0(\bz;\theta)\in\cA$
  \STATE \textbf{for} $l = 0$ to $L$ 
  \STATE \qquad Compute $\hat\nabla_\theta \bbE_{\bz\sim s(\bz;\theta)}\left[\nabla f(q^t(\bz))\right]$ using Equation~\eqref{eq:stoch_grad}
  \STATE \qquad  Compute $s^{l+1}(\bz;\theta)$ from Equation~\ref{eq:proj_grad_step}
  \STATE \textbf{end while}
  \STATE \textbf{return} $s^L(\bz)$
\end{algorithmic}
\caption{stochastic LMO}  \label{algo:LMOstoc}
\end{algorithm} 
\vspace{-3mm}
\subsection{Implementing Line Search}
While one can always perform line search on the original objective, we propose a cheaper alternative which still exhibits the same convergence guarantees. Our alternative can become attractive whenever line search on the $\dkl$ is too expensive computationally.
Let us consider the smoothness quadratic upper bound:
\begin{align*}
f(q^{t+1})\leq \min_{\gamma\in [0,1]}f(q^t) + \gamma \langle s - q^t, \nabla f(q^t)\rangle + \frac{\gamma^2}{2}\Cf
\end{align*}
Instead of performing line search on the original function we compute the stepsize on the quadratic upper bound, which in turns yields a close form solution:
\begin{align*}
\gamma = clip_{[0,1]} \frac{ \langle s - q^t, -\nabla f(q^t)\rangle}{\Cf}
\end{align*}
This quantity can be efficiently estimated via Monte-Carlo sampling as both $s$ and $q^t$ are easy to sample. To sample from $q^t$ one can first sample one of the distribution forming the ensemble and then sample a point from that distribution. 
\subsection{Norm-Corrective Frank-Wolfe}
In this section, we review the norm-corrective Frank-Wolfe~\cite{Locatello:2017tq} which is presented in Algorithm~\ref{algo:normCorrectiveFW}. The main limitation of Algorithm~\ref{algo:FW} is that each iteration uniformly reduces the weights of all the atoms that are active (i.e. 	the densities with non zero weight in the mixture). This is undesirable especially in the variational inference setting where the first approximating densities carries a lot of the information. On the other hand, in the early iterations, suboptimal choices can be made as they are considered optimal by the greedy strategy but lose significance as the optimization proceeds. Therefore, it is useful to selectively update all the weights of the mixtures at the same time. For efficiency reasons, we update all the weights at every iteration but rather than minimizing the $\dkl$ directly we target its quadratic upper bound as we did in the previous section. This results in a quadratic programming problem on the probability simplex (recall that weights sums to one) for which many efficient solutions are known as T is typically small.
\begin{algorithm}[h]
\caption{Norm-Corrective Frank-Wolfe}
\label{algo:normCorrectiveFW}
\begin{algorithmic}[1]
  \STATE \textbf{init} $q^{0} \in \conv(\cA)$, and $\cS:=\{q^{0}\}$
  \STATE \textbf{for} {$t=0\dots T$}
  \STATE \qquad Find $z_t := (\text{Approx-}) \lmo_{\cA}(\nabla f(q^{t}))$
  \STATE \qquad $\cS:=\cS\cup\{ z_t \}$
  \STATE \qquad Let $b := q^{t}-\frac1L \nabla f(q^{t})$
  \STATE \qquad \textit{Variant 0:} Update $q^{t+1}:= \displaystyle\argmin_{z\in \conv(\cS)} \|{z-b}\|_2^2$\\
   \STATE \qquad \textit{Variant 1:} Update $q^{t+1}:= \displaystyle\argmin_{z\in \conv(\cS)} f(z)$\\
  \STATE \qquad \emph{Optional:} Correction of some/all atoms $z_{0\ldots t}$
  \STATE \textbf{end for}
\end{algorithmic}
\end{algorithm}
The name ``norm-corrective'' is used to illustrate that the algorithm relies on a simple quadratic surrogate function (or upper bound on $f$), which only depends on the smoothness constant $L$. This procedure allows for efficient optimization using standard convex solvers. Finding the closest point in norm can typically be performed much more efficiently than solving a general optimization problem on the $\dkl$ over the same domain, which is what the ``fully-corrective'' algorithm variants require in each iteration (Variant 1). Variant 0 of Algorithm~\ref{algo:normCorrectiveFW} is the equivalent of Variant 1 of Algorithm~\ref{algo:FW} where the line search on the quadratic upper bound is performed on all the active atoms rather than just the one added in the current iteration, hence the name corrective.

In \cite{Locatello:2017tq}, the authors showed sublinear convergence of Algorithm~\ref{algo:normCorrectiveFW}. In this work, we show that under some additional assumptions the convergence is actually linear.

\begin{theorem}[\cite{LacosteJulien:2015wj}]\label{thm:FCFWlinear}
Let $\cA \subset \cH$ be a compact set and let $f \colon \cH \to \R$ be both $L$-smooth and $\mu$-strongly convex over the optimization domain. Then, the suboptimality of the iterates of Variant 1 of Algorithm~\ref{algo:normCorrectiveFW} decreases geometrically at each step as:
\begin{equation} \label{eq:linrate}
\varepsilon_{t+1}
\leq \left(1- \beta \right)\varepsilon_{t},
\end{equation}
where $\beta := \delta^2\frac{\mu \text{PWidth}^2}{L\diam(\cA)^2}\in (0,1]$, $\varepsilon_t := f(q^t) - f(q^\star)$ is the suboptimality at step $t$ and $\delta \in (0,1]$ is the relative accuracy parameter of the employed approximate \lmo.
\end{theorem}
In Theorem~\ref{thm:FCFWlinear} we used the notion of pyramidal width:
 \begin{align*}
\mathrm{PWidth}(\cA):= \min_{\substack{\cK\in \faces(\conv(\cA))\\ q\in \cK \\ r\in\cone(\cK-q)\setminus \lbrace \0\rbrace}} PdirW(\cK\cap\cA,r,q).
\end{align*}
For an in depth description of the PWidth, see \cite{LacosteJulien:2015wj}. In the continuous setting, the pyramidal width can be arbitrarily small. For such a reason, quantization of the mean vector is sufficient to ensure that the pyramidal width is bounded away from zero. 
To obtain a linear convergence rate for Variant 0 of Algorithm~\ref{algo:normCorrectiveFW} one needs to upper-bound the number of ``bad steps''. This notion comes from the Pairwise and Away step Frank-Wolfe \cite{LacosteJulien:2015wj}. Let $\bv_t$ be the away vertex $v_t = LMO_\cS(-\nabla f(q^t))$, the exponential decay is not guaranteed when we remove all the weight from $\bv_t$ but $|\cS_t| = |\cS_{t+1}|$. Unfortunately, the tightest known bound  for Variant 0 on the number of good steps is $k(t) \geq t/(3|\cA|!+1)$. The rate of Variant 0 is given in the Appendix.
While this approach is unsatisfactory, the linear convergence of Frank-Wolfe is an active field of research beyond the scope of this paper. In any case, Algorithm~\ref{algo:normCorrectiveFW} is potentially much faster than Algorithm~\ref{algo:FW} at the cost of a greater computation complexity per iteration. 
Furthermore, Algorithm~\ref{algo:FW} is already linearly convergent if the optimum lies in the relative interior of $\conv(A)$ as shown in \cite{guelat1986some}. Therefore, in practice, the norm corrective variant can achieve linear convergence and in general converges faster than Algorithm~\ref{algo:FW}.
\vspace{-2mm}
\paragraph{Discussion} In other words, we showed that with the standard assumptions necessary to show sublinear convergence of FW on the variational inference problem, one can use the full FW framework allowing for potentially globally linearly convergent algorithms. After a quantization of the mean values, the convergence is linear as $\conv(\cA)$ has a finite number of faces. To the best of our knowledge, our results are the first linearly convergent algorithms on the boosting variational inference problem. Furethermore, we identify which assumptions depends on the development of the Frank-Wolfe analysis (bounded pyramidal width for Algorithm~\ref{algo:normCorrectiveFW} or optimum in the relative interior of $\conv(\cA)$ for Algorithm~\ref{algo:FW}). The relation between PWidth and $\diam$ is also known as \textit{condition number} of a set and is related to its eccentricity. Intuitively, a smaller diameter helps the optimization by reducing the size of the search space. On the other hand, in the continuous setting the set $\cS$ can contain atoms forming a very narrow pyramid which in the limit gives vanishing pyramidal width. Unfortunately, computing this constant is challenging and it is known only for few examples, see \cite{LacosteJulien:2015wj}.
\vspace{-2mm}
\section{Experimental Proof of Concept}
\vspace{-2mm}
\paragraph{Synthetic data}
In this section we empirically observe the convergence of Algorithms~\ref{algo:FW} and \ref{algo:normCorrectiveFW} on a toy task verifying that the convergence follows our analysis. In particular, we consider two simple forms for the posterior distribution in 1 dimension, a heavy tailed Cauchy distribution and a mixture of Gaussian distributions. We approximate both distributions using the line search and the fully corrective variants of FW. As expected, even after the rough approximations we performed, the fully corrective perfectly fits the target distribution in a very limited number of iterations. To ensure linear convergence we performed quantization of the mean vectors (stride of $0.0001$). In both examples we used $L=15$ and $L=5$ for line search and the fully corrective respectively. To find the weight in the fully corrective we used standard  semidefinite-quadratic programming (cvx solver). As expected, while being more expensive per iteration, Algorithm~\ref{algo:normCorrectiveFW} converges much faster in terms of number of iterations. Therefore, we showed that linear convergence is achievable using Algorithm~\ref{algo:normCorrectiveFW} while minimizing the $\dkl$. 

 \begin{figure}
\begin{minipage}{0.49\textwidth}
\center\includegraphics[scale=0.25]{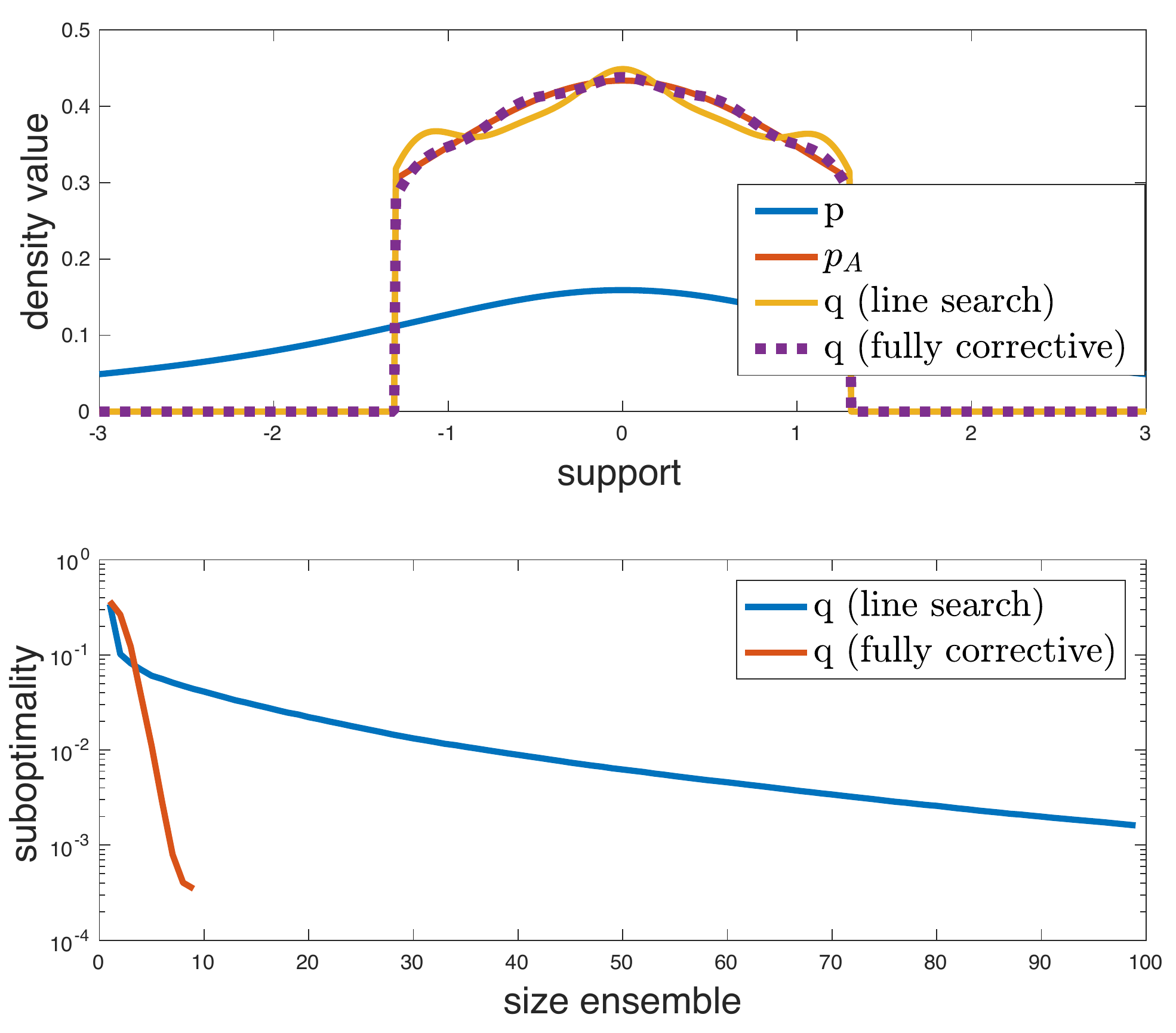}
\captionof{figure}{Convergence of Algorithm~\ref{algo:normCorrectiveFW} compared to \ref{algo:FW} on a truncated cauchy distribution}
\end{minipage}
\hspace{2mm}
\begin{minipage}{0.49\textwidth}
\center\includegraphics[scale=0.25]{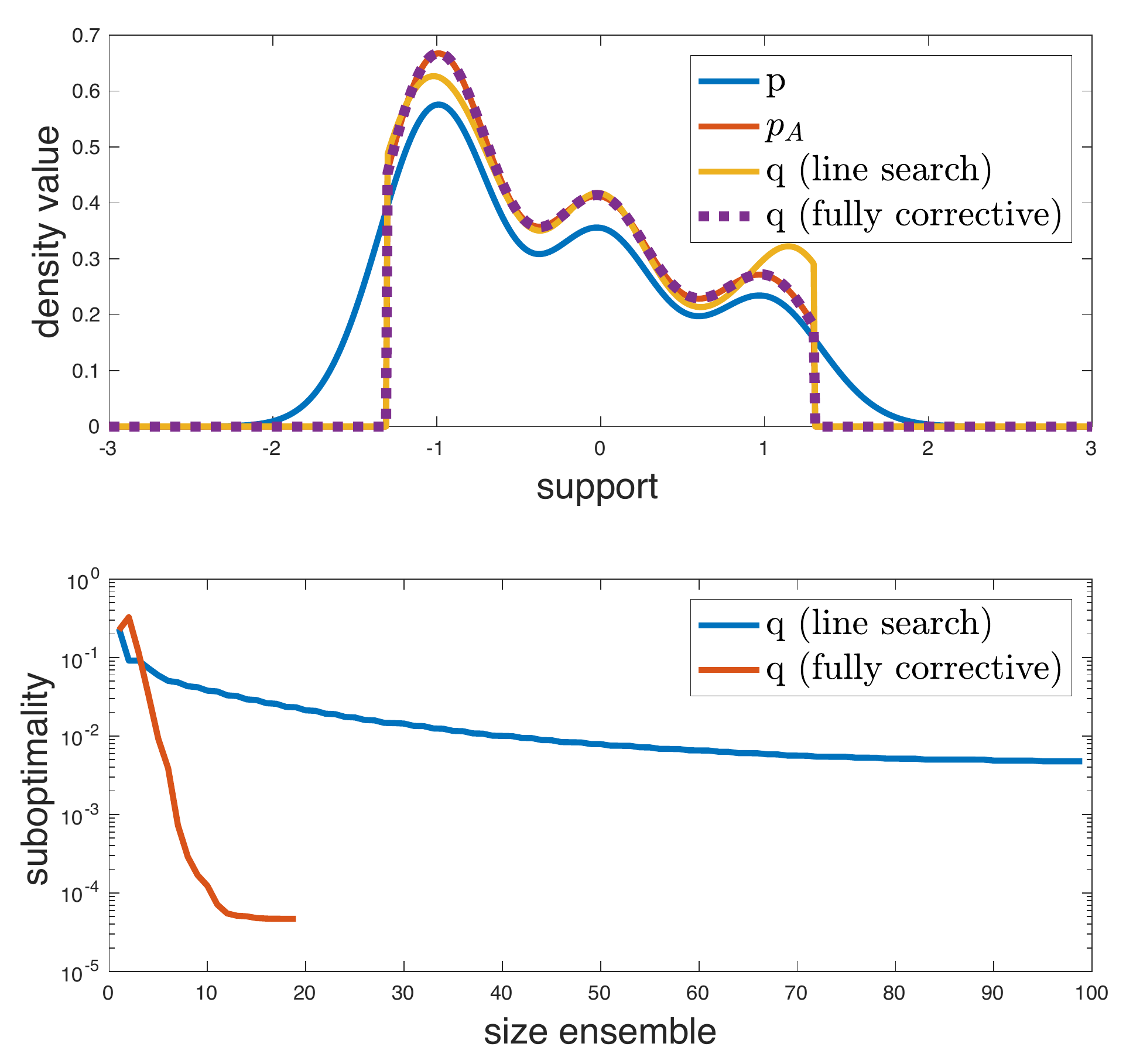}
\captionof{figure}{Convergence of Algorithm~\ref{algo:normCorrectiveFW} compared to \ref{algo:FW} on a truncated mixture of Gaussian distributions}
\end{minipage}
\vspace{-4mm}
\end{figure}
\vspace{-2mm}
\paragraph{Discussion}
In \cite{Guo:2016tg} the authors perform an extensive experimental evaluation showing the remarkable practical performances of Algorithm~\ref{algo:FW}. On the other hand, they do not truncate the Gaussian distributions in the experiments and still observe excellent convergence properties. Note that, provided that the algorithm is initialized well enough, $q/p$ can be bounded away from zero which entails that there exist a finite $L$ which upper bounds the smoothness constant for a fixed and finite number of iterations. As they regularize the LMO with the log of the determinant of the covariance matrix their set $\cA$ has bounded diameter. Therefore, their algorithm is linearly convergent whenever the true posterior is in the relative interior of $\conv(\cA)$ and sublinear otherwise. 
\paragraph{Real Data}
To illustrate the practical utility of the boosting framework, we implement the algorithm for the real world application of predicting whether a chemical is reactive or not (i.e. the response vector $\by$ is binary valued) from its features $\bX$. We use the \textsc{ChemReact} dataset which contains $26733$ chemicals, each with $100$ features. The training data contains $24059$ points, while the rest forms the testing dataset. For the prediction task, we employ the use of Bayesian Logistic Regression with a spherical prior on the regression coefficients $\bw \sim \cN(\mathbf{0}, \bI)$. If $\bx_i \in \R^{100}$ and $y_i \in \{0,1\}$ are the $i^{\text{th}}$ feature vector and response value respectively, then the logistic likelihood function can be written as: 
\begin{align*}
\log p(\by | \bw ; \bX) &:= \sum_i y_i \text{sigmoid}(\bx_i^\top \bw) \\&\qquad+ (1- y_i)[1-  \text{sigmoid}(\bx_i^\top \bw)], 
\end{align*}
where we represent $\bX$ as the feature matrix formed by stacking $\bx_i$, $\by$ is the response vector, and the sigmoid function is $\text{sigmoid}(\alpha) = \frac{1}{1+\exp(-\alpha)}$. It is straightforward to see that the posterior for the above model does not have a closed form expression, nor is it easy to sample from it. Typically, even for such a relatively simple model, MCMC techniques can be prohibitively slow, and so mean field variational inference is often used. 

\begin{figure}
\begin{subfigure}[b]{\textwidth}
\includegraphics[scale=0.28]{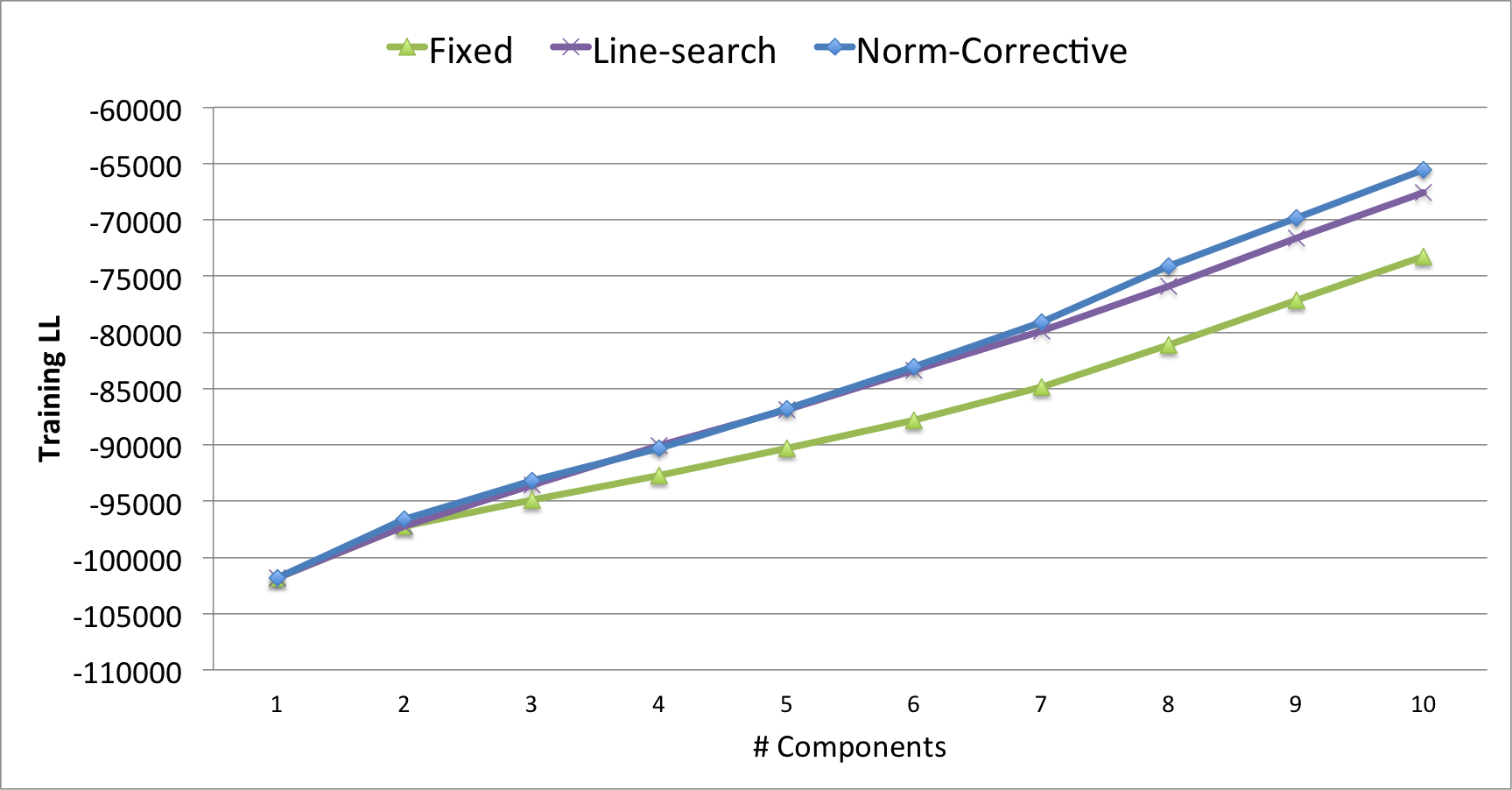}
\end{subfigure}
\begin{subfigure}[b]{\textwidth}
\includegraphics[scale=0.28]{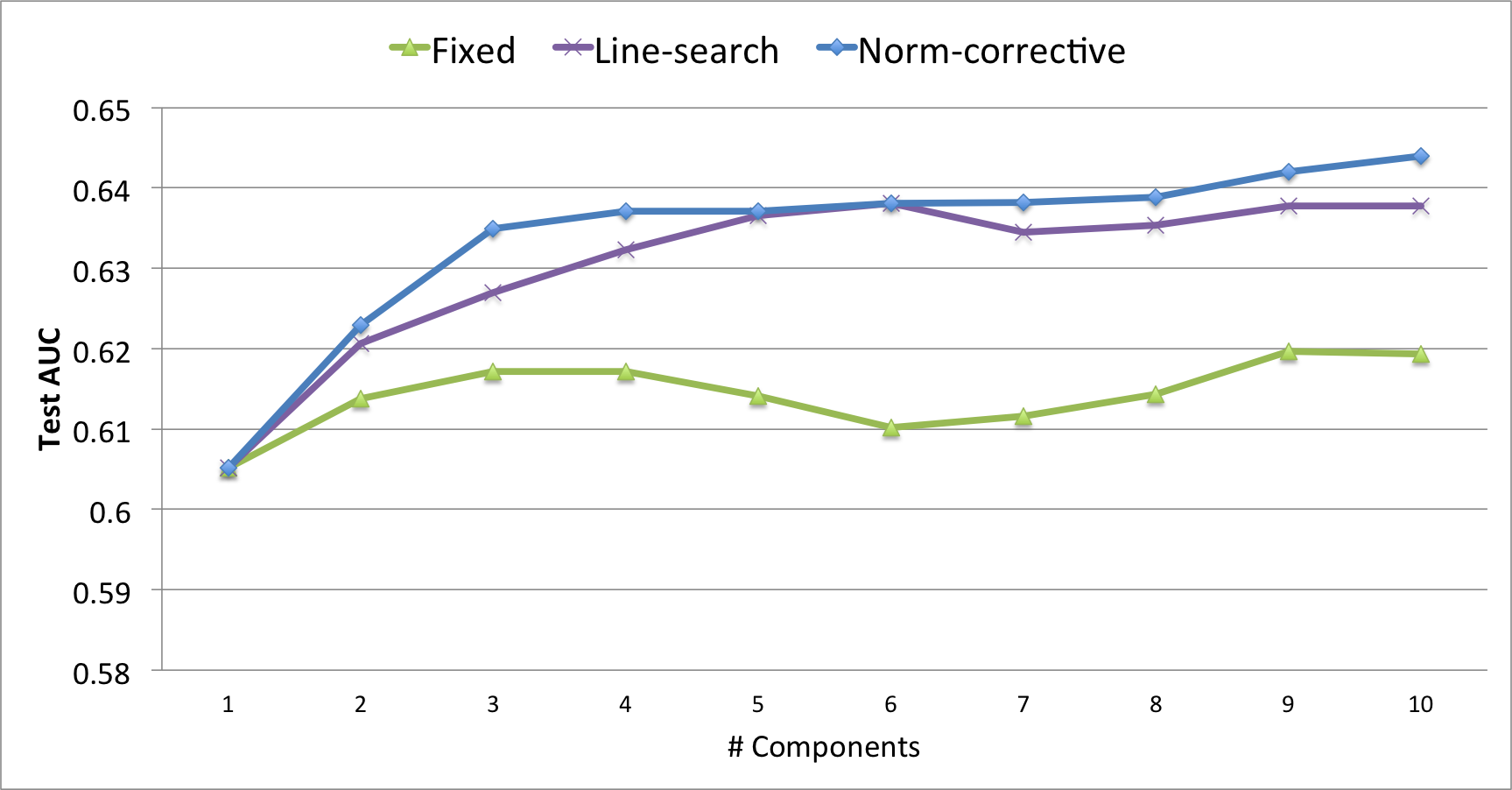}
\end{subfigure}
\caption{Application of different weights optimization techniques for \textit{ChemReact} dataset: norm corrective (Algorithm~\ref{algo:normCorrectiveFW}), line search \cite{Guo:2016tg} and decaying fixed step size (Algorithm~\ref{algo:FW} variant 0)}\label{fig:realdata}
\vspace{-4mm}
\end{figure}

We use the mean field variational inference to initialize our boosting algorithm, and we show that the mixture of gaussians from the mean field field family gives a better training fit and testing accuracy than the vanilla mean field inference. 
We reduce the variance of the gradient estimator with the Rao-Blackwellization \cite{casella1996rao}. To illustrate the importance of the connections with the Frank Wolfe algorithm, we implement three different methods of optimizing over the weights of the mixture. First of all, we implement the line search technique minimizing the original objective already proposed in~\cite{Guo:2016tg}. However, a simpler fixed step size also guarantees convergence as per the FW analysis, and so does the fully corrective step that optimizes over all the previous weights. This is illustrated in Figure~\ref{fig:realdata}. Specifically, we report the training data log-likelihood values to show that the three different techniques offer varying rates of training data fit as expected. The training data fit also translates to the test data accuracy, which we present as the area under the curve (AUC) of the receiver operator characteristic. 

\vspace{-2mm}
\section{Conclusion}
\vspace{-2mm}
We have presented an in-depth theoretical convergence analysis of the boosting variational inference paradigm, delineating explicitly the rates and assumptions that are required for the previously conjectured sublinear and the presented linear convergence rates. \vspace{-2mm}
\paragraph{Acknowledgments:} 
We thank Sahand N. Negahban for the useful discussion.
FL is supported by the Max-Planck ETH Center for Learning Systems. RK is supported by NSF Grant IIS 1421729.

\clearpage
\newpage
{
\setlength{\bibsep}{1.5pt plus .5ex}
\bibliographystyle{plain}
\bibliography{bibliography}
}
\clearpage
\newpage
\appendix
\section{Proof of Lemma \ref{lemma:smooth}}
If $q/p$ is bounded away from zero, $\dkl(q)$ is trivially smooth as it's gradient has bounded norm.

 Viceversa, we need to show that if $\dkl(q)$ is smooth then $q/p$ is bounded away from zero. Since $\dkl(q)$ is smooth, its gradient is absolutely continuous and therefore differentiable almost everywhere with bounded norm. Now, $\nabla \dkl(q) = \log \frac{q}{p}$ and its derivative exists everywhere and is bounded except for a ball around the origin with arbitrary small radius. If by contradiction this ball is in the domain $\cZ$ (i.e. $q/p$ is not bounded away from zero) this set does not have Lesbegue measure zero and thus $\dkl(q)$ is not smooth as its gradient is not absolutely continuous. Note that the $\dkl$ can be locally smooth if $p$ is arbitrarily small in the same region of $q$ and they both decrease equally fast.
 
A sufficient condition is $q$ bounded away from zero everywhere in its support as it would imply $q/p\geq \varepsilon>0$.

\section{Proof of Main Results:}\label{app:proofs}

\begin{reptheorem}{thm:mfBounded}
 The set $\cA$ of non degenerate truncated distributions bounded from above and compact support $\sfA$ is a compact subset of $\cH$.
\begin{proof}
\begin{align*}
\diam(\cA)^2 &= \max_{p,q\in\cA}\|p-q\|^2\\
&\leq \max_{p,q\in\cA}(\|p\| + \|q\|)^2\\
&\leq \max_{q\in\cA}4\|q\|^2
\end{align*}
$q\in\cA$ is defined everywhere in $\sfA$ and is bounded in infinity norm by assumption. The result of the integral is bounded as $\sfA$ is compact. 
In particular:
\begin{align*}
\|q\|^2 &= \int_{\sfA} q(\bz)^2 d\bz\\
&\leq M^2 \int_{\sfA} 1 d\bz
\end{align*}
Now, $\int_{\sfA} 1 d\bz$ is the Lebesgue measure of the set $\sfA$ which is finite as $\sfA$ is compact and non zero as $\sfA$ is full-dimensional.

For truncated gaussian distributions with diagonal covariance matrix we compute a tighter diameter:
\begin{align*}
\|q\|^2 &\leq \int_{\sfA} q(\bz)^2 d\bz\\
&\leq \int_{\sfA} \frac{\cN(\bz,\bmu,\sigma^2I)^2\delta_\sfA(\bz)}{P(\cN(\bz,\bmu,\sigma^2I)\in\cA)^2} d\bz\\
&\leq \int_{\sfZ} \frac{\cN(\bz,\bmu,\sigma^2I)^2}{P(\cN(\bz,\bmu,\sigma^2I)\in\cA)^2} d\bz\\
&\leq \frac{1}{P(\cN(\bz,\bmu,\sigma^2I)\in\cA)^2}\int_{\sfZ} \cN(\bz,\bmu,\sigma^2I)^2  d\bz\\
\end{align*}
and
\begin{align*}
\int_{\sfZ} \cN(\bz,\bmu,\sigma^2I)^2  d\bz = \frac{1}{\sigma^d (2\sqrt{\pi})^{d}}
\end{align*}
Therefore, the maximum norm is:
\begin{align*}
\frac{1}{\sigma_{min}^d(2\sqrt{\pi})^d min_{\bmu\in\sfA}P(\cN(\bz,\bmu,\sigma_{max}I)\in\cA)^2}
\end{align*}
We call $K^2 := min_{\bmu\in\sfA}P(\cN(\bz,\bmu,\sigma_{max}^2I)\in\cA)^2$.
and write:
\begin{align*}
\diam(\cA)^2 \leq \frac{4}{\sigma^d_{min}(2\sqrt{\pi})^dK^2}
\end{align*}
\end{proof}
\end{reptheorem}
\begin{reptheorem}{thm:explicitConstants}
Let the set $\cA$ satisfy A1 and A2. Then, it holds that:
\begin{align*}
\Cf\leq L\diam(\cA)^2\leq 4\frac{M^2} {\epsilon}\cL(\sfA)
\end{align*} 
\begin{proof}
The proof if trivial after showing Theorem~\ref{thm:mfBounded} and recalling that $L = \frac{1}{\epsilon}$
\end{proof}
\end{reptheorem}

\begin{reptheorem}{thm:FWsublinearGaussian}
Let the set $\cA$ of non degenerate truncated Gaussian distribution have compact support $\sfA\in\bbR^d$. Further assume that their means are in $\sfA$ and their covariance matrix before truncation is given by $\sigma^2\bI$ with $\sigma\geq\sigma_{\min}>0$ with $\sigma_{\min}$ being small enough such that $p_{\sfA}\in\conv(\cA)$. Let $\ba$ and $\bb$ be the vertices of the diameter of $\sfA$. Then, the information loss of the Affine Invariant Frank-Wolfe algorithm (Algorithm~\ref{algo:FW}) with some choice of the compact support $\sfA$ converges for $t \geq 0$ as\vspace{-1mm}
\begin{align*}
\dkl(q^t||p) &\leq \frac{4P(\cN(\ba,\sigma_{\min}^2\bI)\in\sfA)}{\sigma_{\min}^{\frac{d}{2}}2^{\frac{d}{2}} K^2}\exp\left(\frac12\frac{\diam(\sfA)^2}{\sigma^2_{\min}}\right)\\&\qquad\frac{1}{\delta^2 t+2} + \frac{2\varepsilon_0}{\delta t+2} -\log p(\bz_{\sfZ\setminus\sfA} = 0)
\end{align*}
where $\varepsilon_0 = \dkl(q^0||p) - \dkl(q^\star||p)$, $\delta \in (0,1]$ is the accuracy parameter of the employed approximate \lmo,  $p$ is the true posterior distribution and $K := min_{\bmu\in\sfA}P(\cN(\bz,\bmu,\sigma_{max}^2I)\in\cA)$. Note that $K$ is bounded away from zero.
\begin{proof}
To show the result we essentially need to compute $\Cf$ for the particular choice in the theorem statement.
Let $\ba,\bb$ be two points $\sfA$ such that the minimum value of any $q\in\cA$ is attained in $\bb$ by a density centered in $\ba$ (wlog). It is trivial to show that these points are the vertices of the diameter of the support $\sfA$.

First of all, recall that:
\begin{align*}
\diam(\cA)^2\leq\frac{4}{\sigma_{\min}^d(2\sqrt{\pi})^d  K^2}
\end{align*}

 The minimal value of any $q\in\cA$ can be computed explicitly as by assumption is reached in $\bb$ by a density centered in $\ba$ with minimal covariance:
\begin{align*}
\epsilon = \frac{\cN(\bb;\ba,\sigma_{\min}^2\bI)}{P(\cN(\ba,\sigma_{\min}^2\bI)\in\sfA)}=\frac{1}{L}.
\end{align*}
Therefore:
\begin{align*}
L\diam(\cA)^2 &\leq \frac{P(\cN(\ba,\sigma_{\min}^2\bI)\in\sfA)}{\cN(\bb;\ba,\sigma_{\min}^2\bI)}\cdot\frac{4}{\sigma_{\min}^d(2\sqrt{\pi})^d  K^2}\\
&= \frac{4P(\cN(\ba,\sigma_{\min}^2\bI)\in\sfA)}{\sigma_{\min}^d(2\sqrt{\pi})^d \cN(\bb;\ba,\sigma_{\min}^2\bI)K^2}\\
&= \frac{4P(\cN(\ba,\sigma_{\min}^2\bI)\in\sfA)}{\sigma_{\min}^{\frac{d}{2}}2^{\frac{d}{2}} K^2}\exp\left(\frac12\frac{\|\ba-\bb\|^2}{\sigma^2_{\min}}\right)\\
&= \frac{4P(\cN(a,\sigma_{\min}^2\bI)\in\sfA)}{\sigma_{\min}^{\frac{d}{2}}2^{\frac{d}{2}} K^2}\exp\left(\frac12\frac{\diam(\sfA)^2}{\sigma^2_{\min}}\right)\\
\end{align*}
As we assumed that $\sigma_{min}$ is small enough to approximate perfectly $p_\sfA$ the proof is concluded.
\end{proof}
\end{reptheorem}

\begin{reptheorem}{thm:FCFWlinear}
Let $\cA \subset \cH$ be a compact set and let $f \colon \cH \to \R$ be both $L$-smooth and $\mu$-strongly convex over the optimization domain.

Then, the suboptimality of the iterates of Variant 0 of Algorithm~\ref{algo:normCorrectiveFW} decreases geometrically at each ``good step'' as:
\begin{equation} 
\varepsilon_{t+1}
\leq \left(1- \beta \right)\varepsilon_{t},
\end{equation}
where $\beta := \delta^2\frac{\mu Pwidth^2}{L\diam(\cA)^2}\in (0,1]$, $\varepsilon_t := f(\bx_t) - f(\bx^\star)$ is the suboptimality at step $t$ and $\delta \in (0,1]$ is the relative accuracy parameter of the employed approximate \lmo.
\begin{proof}
The proof is a trivial extension of the one presented in \cite{LacosteJulien:2015wj}. It only differs in the use of the smoothness upper bound.
Let $v_t = LMO_\cS(-\nabla f(q^t))$
The update of Algorithm~\ref{algo:normCorrectiveFW} yields:
\begin{eqnarray*}
f(q^{t+1}) &= & \min_{q^{t+1}\in\conv(\cS)} f(q^t) + \gamma \langle \nabla f(q^t), q^{t+1} -q^t \rangle   \\
&+& \frac{\gamma^2}{2}L\|q^{t+1} - q^t\|^2 \\
 &\leq & \min_{\gamma \in [0,1]} f(q^t) + \gamma \langle \nabla f(q^t), \tilde{z}_t -v_t \rangle   \\
  &+& \frac{\gamma^2}{2}L\|\tilde{z_t} - v_t\|^2 \\
& =& f(q^t) - \frac{ \left\langle \nabla f(q^t), \tilde{z}_t  -v_t\right\rangle ^2}{2 L \|\tilde{z}_t - v_t\|^2}.
\end{eqnarray*}
This upper bound holds for Algorithm~\ref{algo:normCorrectiveFW} as minimizing the RHS of the first equality coincides with the update of Algorithm~\ref{algo:normCorrectiveFW}. The last equality comes from the assumption that we are performing a good step.
Using $\varepsilon_t = f(q^\star) - f(q^t)$, we can lower bound the error decay as 
\begin{eqnarray}\label{eqproof:linearMPAffineLowerBound}
\varepsilon_{t} - \varepsilon_{t+1} \geq \frac{ \left\langle \nabla f(q^t), \tilde{z}_t  -v_t\right\rangle ^2}{2 L \|\tilde{z_t} - v_t\|^2}.
\end{eqnarray}

The rest of the proof is identical to the one in \cite{LacosteJulien:2015wj} for the Pairwise Frank-Wolfe.
\end{proof}
\end{reptheorem}

\end{document}

%% file: notation.tex
\def\R{{\mathbb{R}}}

\def\to{{\,\rightarrow\,}}

\mathchardef\mhyphen="2D

\newcommand{\vertiii}[1]{{\left\vert\kern-0.25ex\left\vert\kern-0.25ex\left\vert #1
    \right\vert\kern-0.25ex\right\vert\kern-0.25ex\right\vert}}

\newcommand{\vect}[1]{{\boldsymbol{#1}}}

\def\bmu{\vect{\mu}}

\def\ba{{\mathbf{a}}}
\def\bb{{\mathbf{b}}}

\def\bv{{\mathbf{v}}}
\def\bw{{\mathbf{w}}}
\def\bx{{\mathbf{x}}}
\def\by{{\mathbf{y}}}
\def\bz{{\mathbf{z}}}

\def\bI{{\mathbf{I}}}

\def\bX{{\mathbf{X}}}

\def\0{{\mathbf{0}}}

\def\bbE{{\mathbb{E}}}

\def\bbR{{\mathbb{R}}}

\def\cA{\mathcal{A}}

\def\cD{\mathcal{D}}

\def\cH{\mathcal{H}}

\def\cK{\mathcal{K}}
\def\cL{\mathcal{L}}

\def\cN{\mathcal{N}}

\def\cQ{\mathcal{Q}}

\def\cS{\mathcal{S}}

\def\cZ{\mathcal{Z}}

\def\sfA{\mathsf{A}}

\def\sfQ{\mathsf{Q}}

\def\sfZ{\mathsf{Z}}


%% file: theorem_notation.tex
\newtheorem*{rep@theorem}{\rep@title}
\newcommand{\newreptheorem}[2]{%
\newenvironment{rep#1}[1]{%
 \def\rep@title{#2 \ref{##1}}%
 \begin{rep@theorem}}%
 {\end{rep@theorem}}}
\newreptheorem{lemma}{Lemma'}
\newreptheorem{definition}{Definition'}
\newreptheorem{proposition}{Proposition'}
\newreptheorem{theorem}{Theorem'}

\newtheorem{theorem}{Theorem}
\newtheorem{corollary}[theorem]{Corollary}
\newtheorem{lemma}[theorem]{Lemma}

\renewcommand{\text}[1]{{\textnormal{#1}}}

